\documentclass[11pt]{article}
\usepackage{nodalida2005}
\usepackage{latexsym}
\usepackage{graphicx}

\title{Dependency Treebanks: Methods, Annotation Schemes and Tools}

\author{Tuomo Kakkonen\\
  Department of Computer Science\\
  University of Joensuu\\
  P.O. Box 111\\
  FI-80101 Joensuu\\
  {\tt tuomo.kakkonen@cs.joensuu.fi}}

\date{}

\begin{document}
\maketitle
\begin{abstract}
In this paper, current dependency-based treebanks are introduced and analyzed. The methods used for building the resources, the annotation schemes applied, and the tools used (such as POS taggers, parsers and annotation software) are discussed.
\end{abstract}

\section{Introduction}
Annotated data is a crucial resource for developments in computational linguistics and natural language processing. Syntactically annotated corpora, \textit{treebanks}, are needed for developing and evaluating natural language processing applications, as well as for research in empirical linguistics. The choice of annotation type in a treebank usually boils down to two options: the linguistic resource is annotated either according to some constituent or functional structure scheme. As the name treebank suggests, these linguistic resources were first developed in the phrase-structure framework, usually represented as tree-shaped constructions. The first efforts to create such resources started around 30 years ago. The most well-known of such a treebank is the \textit{Penn Treebank} for English~\cite{marcus:93}.

In recent years, there has been a wide interest towards functional annotation of treebanks. In particular, many dependency-based treebanks have been constructed. In addition, grammatical function annotation has been added to some constituent-type treebanks. \textit{Dependency Grammar} formalisms stem from the work of Tesnie\'{e}re~\shortcite{tesnieere:59}. In dependency grammars, only the lexical nodes are recognized, and the phrasal ones are omitted. The lexical nodes are linked with directed binary relations. The most commonly used argument for selecting the dependency format for building a treebank is that the treebank is being created for a language with a relatively free word order. Such treebanks exist \textit{e.g.} for Basque, Czech, German and Turkish. On the other hand, dependency treebanks have been developed for languages such as English, which have been usually seen as languages that can be better represented with constituent formalism. The motivations for using dependency annotation vary from the fact that the type of structure is the one needed by many, if not most, applications to the fact that it offers a proper interface between syntactic and semantic representation. Furthermore, dependency structures can be automatically converted into phrase structures if needed~\cite{lin:95,xia:01}, although not always with 100\% accuracy. The \textit{TIGER Treebank} of German, a free word order language, with 50,000 sentences is an example of a treebank with both phrase structure and dependency annotations~\cite{brants:02}. 

The aim of this paper is to answer the following questions about the current state-of-art in dependency treebanking:
\begin{itemize} {\leftmargin 0.5cm}
	\item What kinds of texts do the treebanks consist of?
	\item What types of annotation schemes and formats are applied? 
	\item What kinds of annotation methods and tools are used for creating the treebanks?
	\item What kinds of functions do the annotation tools for creating the treebanks have?
\end{itemize}

We start by introducing the existing dependency-based treebanks (Section 2). In Section 3, the status and state-of-art in dependency treebanking is summarized and analyzed. Finally in Section 4, we conclude the findings. 

\section{Existing dependency treebanks}

\subsection{Introduction}
Several kinds of resources and tools are needed for constructing a treebank: \textit{annotation guidelines} state the conventions that guide the annotators throughout their work, a software tool is needed to aid the annotation work, and in the case of semi-automated treebank construction, a \textit{part-of-speech (POS) tagger}, \textit{morphological analyzer} and/or a \textit{syntactic parser} are also needed. Building trees manually is a very slow and error-prone process. The most commonly used method for developing a treebank is a combination of automatic and manual processing, but the practical method of implementation varies considerably. There are some treebanks that have been annotated completely manually, but with taggers and parsers available to automate some of the work such a method is rarely employed in state-of-the-art treebanking.

\subsection{The Treebanks}

\subsubsection{Prague Dependency Treebank}
The largest of the existing dependency treebanks (around 90,000 sentences), the \textit{Prague Dependency Treebank} for Czech, is annotated in layered structure annotation, consisting of three levels: morphological, analytical (syntax), and tectogrammatical (semantics)~\cite{boh:03}. The data consist of newspaper articles on diverse topics (\textit{e.g.} politics, sports, culture) and texts from popular science magazines, selected from the \textit{Czech National Corpus}. There are 3,030 morphological tags in the morphological tagset~\cite{hajic:98}. The syntactic annotation comprises of 23 dependency types.

The annotation for the levels was done separately, by different groups of annotators. The morphological tagging was performed by two human annotators selecting the appropriate tag from a list proposed by a tagging system. Third annotator then resolved any differences between the two annotations. The syntactic annotation was at first done completely manually, only by the aid of ambiguous morphological tags and a graphical user interface. Later, some functions for automatically assigning part of the tags were implemented. After some 19,000 sentences were annotated, \textit{Collins lexicalized stochastic parser}~\cite{collins:99} was trained with the data, and was capable of assigning 80\% of the dependencies correct. At that stage, the work of the annotator changed from building the trees from scratch to checking and correcting the parses assigned by the parser, except for the analytical functions, which still had to be assigned manually. The details related to the tectogrammatical level are omitted here. Figure 1 illustrates an example of morphological and analytical levels of annotation.

\begin{figure*}
\begin{center}
\small{\textit{
$<$f cap$>$Do$<$l$>$do$<$t$>$RR--2----------$<$A$>$AuxP$<$r$>$1$<$g$>$7\linebreak
$<$f num$>$15$<$l$>$15$<$t$>$C=-------------$<$A$>$Atr $<$r$>$2$<$g$>$4\linebreak
$<$d$>$.$<$l$>$.$<$t$>$Z:-------------$<$A$>$AuxG$<$r$>$3$<$g$>$2\linebreak
$<$f$>$kv\v{e}tna$<$l$>$kv\v{e}ten$<$t$>$NNIS2-----A----$<$A$>$Adv$<$r$>$4$<$g$>$1\linebreak
$<$f$>$budou$<$l$>$b\'{y}t$<$t$>$VB-P---3F-AA---$<$A$>$AuxV$<$r$>$5$<$g$>$7\linebreak
$<$f$>$cestuj\'{i}c\'{i}$<$l$>$cestuj\'{i}c\'{i}$<$t$>$NNMP1-----A----$<$A$>$Sb $<$r$>$6$<$g$>$7\linebreak
$<$f$>$platit$<$l$>$platit$<$t$>$Vf--------A----$<$A$>$Pred$<$r$>$7$<$g$>$0\linebreak
$<$f$>$dosud$<$l$>$dosud$<$t$>$Db-------------$<$A$>$Adv$<$r$>$8$<$g$>$9\linebreak
$<$f$>$platn\'{y}m$<$l$>$platn\'{y}$<$t$>$AAIS7----1A----$<$A$>$Atr$<$r$>$9$<$g$>$10\linebreak
$<$f$>$zp\o{u}sobem$<$l$>$zp\.{u}sob$<$t$>$NNIS7-----A----$<$A$>$Adv$<$r$>$10$<$g$>$7\linebreak
$<$d$>$.$<$l$>$.$<$t$>$Z:-------------$<$A$>$AuxK$<$r$>$11$<$g$>$0\linebreak
}
}
\caption{A morphologically and analytically annotated sentence from the Prague Dependency Treebank.}
\label{fig:prague}
\end{center}
\end{figure*}

There are other treebank projects using the framework developed for the Prague Dependency Treebank. \textit{Prague Arabic Dependency Treebank}~\cite{hajic:04}, consisting of around 49,000 tokens of newswire texts from \textit{Arabic Gigaword} and \textit{Penn Arabic Treebank}, is a treebank of Modern Standard Arabic. The \textit{Slovene Dependency Treebank} consists of around 500 annotated sentences obtained from the \textit{MULTEXT-East Corpus}~\cite{slo:1,slo:2}.

\subsubsection{TIGER Treebank}
The \textit{TIGER Treebank} of German~\cite{brants:02} was developed based on the \textit{NEGRA Corpus}~\cite{skut:98} and consists of complete articles covering diverse topics collected from a German newspaper. The treebank has around 50,000 sentences. The syntactic annotation combining both phrase-structure and dependency representations is organized as follows: phrase categories are marked in non-terminals, POS information in terminals and syntactic functions in the edges. The syntactic annotation is rather simple and flat in order to reduce the amount of attachment ambiguities. An interesting feature in the treebank is that a MySQL database is used for storing the annotations, from where they can be exported into \textit{NEGRA Export} and \textit{TIGER-XML} file formats, which makes it usable and exchangeable with a range of tools.

The annotation tool \textit{Annotate} with two methods, interactive and \textit{Lexical-Functional Grammar} (LFG) parsing, was employed in creating the treebank. \textit{LFG parsing} is a typical semi-automated annotation method, comprising of processing the input texts by a parser and a human annotator disambiguating and correcting the output. In the case of TIGER Treebank, a broad coverage LFG parser is used, producing the constituent and functional structures for the sentences. As almost every sentence is left with unresolved ambiguities, a human annotator is needed to select the correct ones from the set of possible parses. As each sentence of the corpus has several thousands of possible LFG representations, a mechanism for automatically reducing the number of parses is applied, dropping the number of parses represented to the human annotator to 17 on average. \textit{Interactive annotation} is also a type of semi-automated annotation, but in contrast to human post-editing, the method makes the parser and the annotator to interact. First, the parser annotates a small part of the sentence and the annotator either accepts or rejects it based on visual inspection. The process is repeated until the sentence is annotated completely.

\subsubsection{Arboretum, L'Arboratoire, Arborest and Floresta Sint\'{a}(c)tica}
\textit{Arboretum} of Danish~\cite{bick:03}, \textit{L'Arboratoire} of French and \textit{Floresta Sint\'{a}(c)tica} of Portuguese~\cite{alfonso:02}, and \textit{Arborest} of Estonian~\cite{bick:05} are "sibling" treebanks, {Arboretum} being the "oldest sister". The treebanks are hybrids with both constituent and dependency annotation organized into two separate levels. The levels share the same morphological tagset. The dependency annotation is based on the \textit{Constraint Grammar} (CG)~\cite{karlsson:90} and consists of 28 dependency types. For creating each of the four treebanks, a CG-based morphological analyzer and parser was applied. The annotation process consisted of CG parsing of the texts followed by conversion to constituent format, and manual checking of the structures.

Arboretum has around 21,600 sentences annotated with dependency tags, and of those, 12,000 sentences have also been marked with constituent structures~\cite{bick:03,bick:05b}. The annotation is in both TIGER-XML and PENN export formats. Floresta Sint\'{a}(c)tica consists of around 9,500 manually checked (version 6.8, October 15th, 2005) and around 41,000 fully automatically annotated sentences obtained from a corpus of newspaper Portuguese~\cite{alfonso:02}. \textit{Arborest} of Estonian consists of 149 sentences from newspaper articles~\cite{bick:05}. The morphosyntactic and CG-based surface syntactic annotation are obtained from an existing corpus, which is converted semi-automatically to Arboretum-style format.

\subsubsection{The Dependency Treebank for Russian}
The \textit{Dependency Treebank for Russian} is based on the \textit{Uppsala University Corpus}~\cite{lonngren:93}. The texts are collected from contemporary Russian prose, newspapers, and magazines~\cite{boguslavsky:00,boguslavsky:02}. The treebank has about 12,000 annotated sentences. The annotation scheme is XML-based and compatible with \textit{Text Encoding for Interchange} (TEI), except for some added elements. It consists of 78 syntactic relations, divided into six subgroups, such as attributive, quantitative, and coordinative. The annotation is layered, in the sense that the levels of annotation are independent and can be extracted or processed independently. 

The creation of the treebank started by processing the texts with a morphological analyzer and a syntactic parser, \textit{ETAP}~\cite{apresjan:92}, and was followed by post-editing by human annotators. Two tools are available for the annotator: a \textit{sentence boundary markup tool} and \textit{post-editor}. The post-editor offers the annotator functions for building, editing, and managing the annotations. The editor has a special split-and-run mode, used when the parsers fails to produce a parse or creates a parse with a high number of errors. In the mode the user can pre-chunk the sentence into smaller pieces to be input to the parser. The parsed chucks can be linked by the annotator, thus producing a full parse for the sentence. The tool also provides the annotator with the possibility to mark the annotation of any word or sentence as doubtful, in order to remind at the need for a later revision.

\subsubsection{Alpino}
The \textit{Alpino Treebank} of Dutch, consisting of 6,000 sentences, is targeted mainly at parser evaluation and comprises of newspaper articles~\cite{vanderbeek:02}. The annotation scheme is taken from the \textit{CGN Corpus} of spoken Dutch~\cite{oostdijk:00} and the annotation guidelines are based on the TIGER Treebank's guidelines.

The annotation process in the Alpino Treebank starts with applying a parser based on \textit{Head-Driven Phrase Structure Grammar} (HPSG)~\cite{pollard:94} and is followed by a manual selection of the correct parse trees. An interactive lexical analyzer and a constituent marker tools are employed to restrict the number of possible parses. The \textit{interactive lexical analyzer} tool lets the user to mark each word in a sentence belonging to correct, good, or bad categories. 'Correct' denotes that the parse includes the lexical entry in question, 'good' that the parse may include the entry, and 'bad' that the entry is incorrect. The parser uses this manually reduced set of entries, thus generating a smaller set of possible parses. With the constituent marker tool, the annotator can mark constituents and their types to sentences, thus aiding the parser. 

The selection of the correct parse is done by the help of a \textit{parse selection tool}, which calculates \textit{maximal discriminants} to help the annotator. There are three types of discriminants. Maximal discriminants are sets of shortest dependency paths encoding differences between parses, lexical discriminants represent ambiguities resulting from lexical analysis, and constituent discriminants group words to constituents without specifying the type of the constituent. The annotator marks each of the maximal discriminants as good or bad, and the tool narrows down the number of possible parses based on the information. If the parse resulting from the selection is not correct, it can be edited by a parse editor tool.

\subsubsection{The Danish Dependency Treebank}
The annotation of the \textit{Danish Dependency Treebank} is based on \textit{Discountinuous Grammar}, which is a formalism closely related to \textit{Word Grammar}~\cite{kromann:03}. The treebank consists of 5,540 sentences covering a wide range of topics. The morphosyntactic annotation is obtained from the \textit{PAROLE Corpus}~\cite{keson:2005}, thus no morphological analyzer or POS tagger is applied. The dependency links are marked manually by using a command-line interface with a graphical parse view. A parser for automatically assigning the dependency links is under development.

\subsubsection{METU-Sabanci Turkish Treebank}
\begin{figure*}
	\centering
		\includegraphics[scale=0.7]{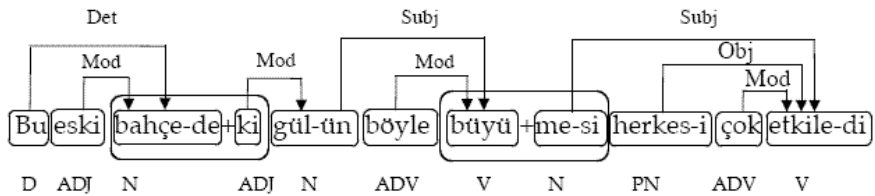}
	\caption{A sample sentence from the METU-Sabanci Treebank.}
	\label{fig:turk}
\end{figure*}

Morphologically and syntactically annotated \textit{Turkish Treebank} consists of 5,000 sentences obtained from the~\textit{METU Turkish Corpus}~\cite{atalay:03} covering 16 main genres of present-day written Turkish~\cite{oflazer:03}. The annotation is presented in a format that is in conformance with the~\textit{XML-based Corpus Encoding Standard} (XCES) format~\cite{ide:03}. Due to morphological complexity of Turkish, morphological information is not encoded with a fixed set of tags, but as sequences of \textit{inflectional groups} (IGs). An IG is a sequence of inflectional morphemes, divided by derivation boundaries. The dependencies between IGs are annotated with the following 10 link types: subject, object, modifier, possessor, classifier, determiner, dative adjunct, locative adjunct, ablative adjunct, and instrumental adjunct. Figure~\ref{fig:turk} illustrates a sample annotated sentence from the treebank.

The annotation, directed by the guidelines, is done in a semi-automated fashion, although relatively lot of manual work remains. First, a morphological analyzer based on the two-level morphology model~\cite{oflazer:94} is applied to the texts. The morphologically analyzed and preprocessed text is input to an annotation tool. The tagging process requires two steps: morphological disambiguation and dependency tagging. The annotator selects the correct tag from the list of tags proposed by the morphological analyzer. After the whole sentence has been disambiguated, dependency links are specified manually. The annotators can also add notes and modify the list of dependency link types.

\subsubsection{The Basque Dependency Treebank}
The \textit{Basque Dependency Treebank}~\cite{aduriz:03} consists of 3,000 manually annotated sentences from newspaper articles. The syntactic tags are organized as a hierarchy. The annotation is done by aid of an annotation tool, with tree visualization and automatic tag syntax checking capabilities.

\subsubsection{The Turin University Treebank}
The \textit{Turin University Treebank} for Italian consisting of 1,500 sentences is divided into four sub-corpora~\cite{lesmo:02,bosco:00,bosco:03}. The majority of texts is from civil law code and newspaper articles. The annotation format is based on the \textit{Augmented Relational Structure} (ARS). The POS tagset consists of 16 categories and 51 subcategories. There are around 200 dependency types, organized as a taxonomy of five levels. The scheme provides the annotator with the possibility of marking a relation as \textit{under-specified} if a correct relation type cannot be determined.

The annotation process consists of automatic tokenization, morphological analysis and POS disambiguation, followed by syntactic parsing~\cite{lesmo:02}. The annotator can interact with the parser through a graphical interface, in a similar way to the interactive method in the TIGER Treebank. The annotator can either accept or reject the suggested tags for each word in the sentence after which the parser proceeds to the next word~\cite{bosco:00}.

\subsubsection{The Dependency Treebank of English}
The \textit{Dependency Treebank of English} consists of dialogues between a travel agent and customers~\cite{rambow:02}, and is the only dependency treebank with spoken language annotation. The treebank has about 13,000 words. The annotation is a direct representation of lexical predicate-argument structure, thus arguments and adjuncts are dependents of their predicates and all function words are attached to their lexical heads. The annotation is done at a single, syntactic level, without surface representation for surface syntax, the aim being to keep the annotation process as simple as possible. Figure 3 shows an example of an annotated sentence~\cite{rambow:02}. 

\begin{figure*}
	\centering
		\includegraphics[scale=0.8]{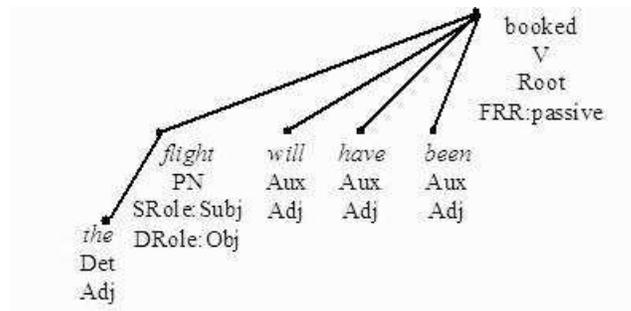}
	\label{fig:eng}
	\caption{The sentence "The flight will have been booked" from the English treebank. The words are marked with the word form (first line), the POS (second line), and the surface role (third line). In addition, node 'flight' is marked with a deep role (DRole) and the root node as passive in the FRR feature, not set in any other nodes.}
\end{figure*}

The trained annotators have access to an on-line manual and work off the transcribed speech without access to the speech files. The dialogs are parsed with a dependency parser, the ~\textit{Supertagger and Lightweight Dependency Analyzer}~\cite{bangalore:99}. The annotators correct the output of the parser using a graphical tool, the one developed by Prague Dependency Treebank project. In addition to the standard tag editing options, annotators can add comments. After the editing is done, the sentence is automatically checked for inconsistencies, such as the difference in surface and deep roles or prepositions missing objects \textit{etc.}

\subsubsection{DEPBANK}
As the name suggests, the \textit{PARC 700 Dependency Bank} (DEPBANK)~\cite{king:03} consists of 700 annotated sentences from the \textit{Penn Wall Street Journal Treebank}~\cite{marcus:94}. There are 19 grammatical relation types (\textit{e.g.} subject, object, modifier) and 37 feature types (\textit{e.g.} number (pl/sg), passive (+/-), tense (future/past/present)) in the annotation scheme. 

The annotation process is semi-automatic, consisting of parsing by broad-coverage LFG, converting the parses to the DEPBANK format and manually checking and correcting the resulting structures. The annotations are checked by a tool that looks \textit{e.g.} for the correctness of header information and the syntax of the annotation, and inconsistencies in feature names. The checking tool helps in two different ways: first, when the annotator makes corrections to the parsed structure, it makes sure that no errors were added, and second, the tool can detect erroneous parses and note that to the annotator. 

\section{Analysis}

Table ~\ref{table:treebanks} summarizes some key properties of the existing dependency treebanks. The size of the treebanks is usually quite limited, ranging from few hundreds to 90,000 sentences. This is partly due to the fact that even the most long-lived of the dependency treebank projects, the Prague Dependency Treebank, was started less than 10 years ago. The treebank producers have in most cases aimed at creating a multipurpose resource for evaluating and developing NLP systems and for studies in theoretical linguistics. Some are built for specific purposes, \textit{e.g.} the Alpino Treebank of Dutch is mainly for parser evaluation. Most of the dependency treebanks consist of written text; to our knowledge there is only one that is based on a collection of spoken utterances. The written texts are most commonly obtained from newspaper articles, and in the cases of \textit{e.g.} Czech, German, Russian, Turkish, Danish, and Dutch treebanks from an existing corpus. Annotation usually consists of POS and morphological levels accompanied by dependency-based syntactic annotation. In the case of the Prague Dependency Treebank a higher, semantic layer of annotation is also included.

\begin{table*}
\caption{Comparison of dependency treebanks. \small{(*Due to limited number of pages not all the treebanks in the Arboretum "family" are included in the table. **Information of number of utterances was not available. M=manual, SA=semi-automatic, TB=treebank})}
\label{table:treebanks}
\begin{tabular}{|p{1.1cm}|p{1.1cm}|p{1.4cm}|p{1.1cm}|p{2.7cm}|p{2.7cm}|p{2.7cm}|}
\hline
$Name$&
$Lan-guage$&
$Genre$&
$Size \newline (sent.)$&
$Annotation \newline methods$&
$Autom. tools$&
$Supported \newline formats$\\
\hline
\hline
\textit{Prague Dep. TB}&
Czech&
Newsp., science mag.&
90,000&
M/SA&
Lexicalized stochastic parser (Collins)&
FS, CSTS SGML, Annotation Graphs XML\\
\hline
\textit{TIGER TB}&
Ger-man&
Newsp.&
50,000&
Post-editing \& interactive&
Probabilistic/ LFG parser&
TIGER-XML \& NEGRA export\\
\hline
\textit{Arbore-tum \& co.*}&
4 lang.&
Mostly newsp.&
21,600 (Ar)\newline9,500 (Flo)&
Dep. to const. conversion, M checking&
CG-based parser for each language&
TIGER-XML \& PENN export (Ar.)\\
\hline
\textit{Dep. TB for Russian}&
Rus-sian&
Fiction, newsp., scientific&
12,000&
SA&
Morph. analyzer \& a parser&
XML-based TEI-compatible\\
\hline
\textit{Alpino}&
Dutch&
Newsp.&
6,000&
M disambig. aided by parse selection tool&
HPSG-based Alpino parser&
Own XML-based\\
\hline
\textit{Danish Dep. TB}&
Dan-ish&
Range of topics \& genres	&
5,540&
Morphosyn. annotation obtained from a corpus, M dep. marking&
-&
PAROLE-DK with additions, TIGER-XML\\
\hline
\textit{METU-Saba-nci TB}&
Turk-ish&
16 genres&
5,000&
M disambiguation \& M dependency marking &
Morph. analyzer based on XEROX FST&
XML-based XCES compatible\\
\hline
\textit{Basque TB}&
Basque&
Newsp.&
3,000&
M, automatic checking&
-&
XML-based TEI-compatible\\
\hline
\textit{TUT}&
Italian&
Mainly newsp. \& civil law&
1,500&
M checking of parser \& morph. analyzer output&
Morph. analyzer, rule-based tagger and a parser&
Own ASCII-based\\
\hline
\textit{Dep. TB of English}&
Eng-lish&
Spoken, travel agent dial.&
13,000 words\newline**&
M correction of parser output \& autom. checking of inconsistencies&
Supertagger \& Lightweight Dep. Analyzer&
FS\\
\hline
\textit{DEP-BANK}&
Eng-lish&
Financial newsp.&
700&
M checking \& correction, autom. consistency checking &
LFG parser, checking tool&
Own ASCII-based\\
\hline
\end{tabular}
\end{table*}

The definition of the annotation schema is always a trade-off between the accuracy of the representation, data coverage and cost of treebank development~\cite{bosco:00,bosco:03}. The selection of the tagsets for annotation is critical. Using a large variety of tags provides a high accuracy and specialization in the description, but makes the annotators' work even more time-consuming. In addition, for some applications, such as training of statistical parsers, highly specific annotation easily leads into sparsity problem. On the other hand, if annotation is done at a highly general level the annotation process is faster, but naturally lot of information is lost. The TUT and Basque treebanks try to tackle the problem by organizing the set of grammatical relations into hierarchical taxonomy. Also the choice of type of application for the treebank may affect the annotation choices. A treebank for evaluation allows for some remaining ambiguities but no errors, while the opposite may be true for a treebank for training~\cite{abe:03}. In annotation consisting of multiple levels clear separation between the levels is a concern. The format of the annotation is also directed by the specific language that the treebank is being developed for. The format must be suited for representing the structures of the language. For example, in the METU-Sabanci Treebank a special type of morphological annotation scheme was introduced due to the complexity of Turkish morphology. 

Semi-automated creation combining parsing and human checker is the state-of-art annotation method. None of the dependency treebanks are created completely manually; at least an annotation tool capable of visualizing the structures is used by each of the projects. Obviously, the reason that there aren't any fully automatically created dependency treebanks is the fact there are no parsers of free text capable of producing error-free parses. 

The most common way of combining the human and machine labor is to let the human work as a post-checker of the parser's output. Albeit most straight-forward to implement, the method has some pitfalls. First, starting annotation with parsing can lead to high number of unresolved ambiguities, making the selection of the correct parse a time-consuming task. Thus, a parser applied for treebank building should perform at least some disambiguation to ease the burden of annotators. Second, the work of post-checker is mechanic and there is a risk that the checker just accept the parser's suggestions, without a rigorous inspection. A solution followed \textit{e.g.} by the both treebanks for English and the Basque treebank is to apply a post-checking tool to the created structures before accepting them. Some variants of semi-automated annotation exist: the TIGER, TUT, Alpino, and the Russian Treebanks apply a method where the parser and the annotator can interact. The advantage of the method is that when the errors by the parser are corrected by the human at the lower levels, they do not multiply into the higher levels, thus making it more probable that the parser produces a correct parse. In some annotation tools, such as the tools of the Russian, the English Dependency treebanks, the annotator is provided with the possibility of adding comments to annotation, easing the further inspection of doubtful structures. In the annotation tool of the TUT Treebank, a special type relation can be assigned to mark doubtful annotations.

Although more collaboration has emerged between treebank projects in recent years, the main problem with current treebanks in regards to their use and distribution is the fact that instead of reusing existing formats, new ones have been developed. Furthermore, the schemes have often been designed from theory and even application-specific viewpoints, and consequently, undermine the possibility for reuse. Considering the high costs of treebank development (for example in the case of the Prague Dependency Treebank estimated USD600,000~\cite{boh:03}), reusability of tools and formats should have a high priority. In addition to the difficulties for reuse, creating a treebank-specific representation format requires developing a new set of tools for creating, maintaining and searching the treebank. Yet, the existence of exchange formats such as XCES~\cite{ide:03} and TIGER-XML~\cite{mengel:00} would allow multipurpose tools to be created and used. 

\section{Conclusion}
We have introduced the state-of-art in dependency treebanking and discussed the main characteristics of current treebanks. The findings reported in the paper will be used in designing and constructing an annotation tool for dependency treebanks and constructing a treebank for Finnish for syntactic parser evaluation purposes. The choice of dependency format for a treebank for evaluating syntactic parser of Finnish is self-evident, Finnish being a language with relatively free word order and all parsers for the language working in the dependency-framework. The annotation format will be some of the existing XML-based formats, allowing existing tools to be applied for searching and editing the treebank. 

The findings reported in this paper indicate that the following key properties must be implemented into the annotation tool for creating the treebank for Finnish: 

\begin{itemize} {\leftmargin 0.5cm}
	\item An interface to a morphological analyzer and parser for constructing the initial trees. Several parsers can be applied in parallel to offer the annotator a possibility to compare the outputs.
	\item Support for an existing XML annotation format. Using an existing format will make the system more reusable. XML-based formats offer good syntax-checking capabilities.
	\item An inconsistency checker. The annotated sentences to be saved will be checked against errors in tags and annotation format. In addition to XML-based validation of the syntax of the annotation, the inconsistency checker will inform the annotator about several other types of mistakes. The POS and morphological tags will be checked to find any mismatching combinations. A missing main verb, a fragmented, incomplete parse etc. will be indicated to the user.		
	\item A comment tool. The annotator will be able to add comments to the annotations to aid later revision.
	\item Menu-based tagging. In order to minimize errors, instead of typing the tags, the annotator will only be able to set tags by selecting them from predefined lists.
\end{itemize}

\section{Acknowledgments}
The research reported in this work has been supported by the European Union under a Marie Curie Host Fellowship for Early Stage Researchers Training at MULTILINGUA, Bergen, Norway and MirrorWolf project funded by the National Technology Agency of Finland (TEKES). 

\label{sec:citations}

\bibliographystyle{acl}
\bibliography{nodalida2005}

\end{document}